# Churn analysis using deep convolutional neural networks and autoencoders

Artit Wangperawong[a,b,*]
Cyrille Brun[c]
Dr. Olav Laudy[c]
Rujikorn Pavasuthipaisit[a]

[a] True Corporation Public Company Limited, Bangkok 10310, Thailand
[b] Department of Electrical Engineering, Faculty of Engineering, King Mongkut's University of Technology Thonburi, Bangkok 10140, Thailand
[c] International Business Machines Corporation, Singapore 486048

[*] To whom correspondence should be addressed; Email: artitw@gmail.com

**Customer temporal behavioral data was represented as images in order to perform churn prediction by leveraging deep learning architectures prominent in image classification. Supervised learning was performed on labeled data of over 6 million customers using deep convolutional neural networks, which achieved an AUC of 0.743 on the test dataset using no more than 12 temporal features for each customer. Unsupervised learning was conducted using autoencoders to better understand the reasons for customer churn. Images that maximally activate the hidden units of an autoencoder trained with churned customers reveal ample opportunities for action to be taken to prevent churn among strong data, no voice users.**

Deep learning by convolutional neural networks (CNNs) has demonstrated superior performance in many image processing tasks [1,2,3]. In order to leverage such advances to predict churn and take pro-active measures to prevent it, we represent customers as images. Specifically, we construct a 2-dimensional array of normalized pixels where each row is for each day and each column is for each type of behavior tracked (Fig. 1). The type of behavior can include data usage, top up amount, top up frequency, voice calls, voice minutes, SMS messages, etc. In the training and testing data, each image is also accompanied by its label – 1 for churned and 0 for not churned. For this analysis, we examine prepaid customers in particular.

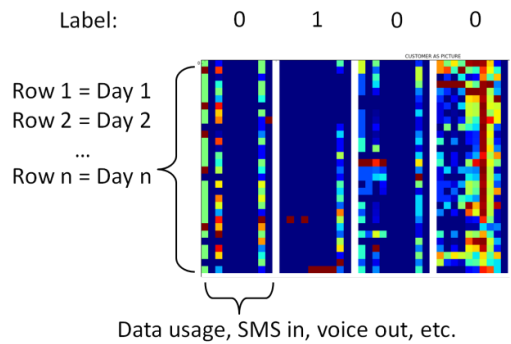

**Figure 1.** Four customers represented as images based on their usage behavior (columns) over a 30-day period (rows). Pixel values increase from blue to red.

In order to determine the labels and the specific dates for the image, we first define *churn*, *last call* and the *predictor window* according to each customer's lifetime-line (LTL). This is best understood by viewing Fig. 2 from right to left. The first item is the churn assessment window, which we have chosen to be 30 days. If the customer registers any activity within these 30 days, we label them with 0 for active/not-churned. In Fig. 2, a green circle demarks this label for the first, top-most customer LTL. If the customer has no activity in this time frame, then we label them as 1 for churned. These are the second and third LTLs in Fig. 2.

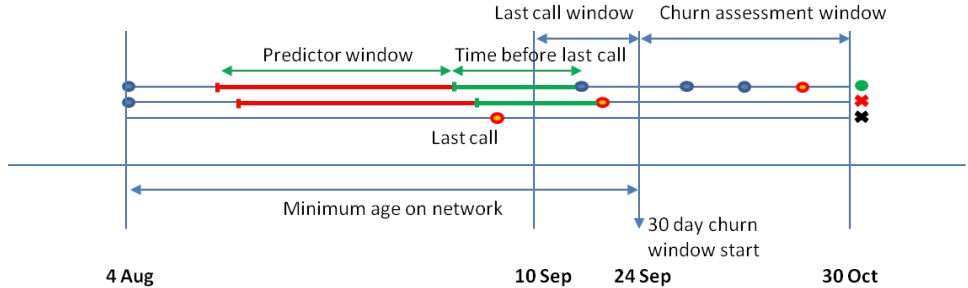

**Figure 2.** Three distinct customer LTLs used to define churn, last call and the 30-day predictor window.

Next, we define the last call, which is the latest call occurring in the 14-day last call window of Fig. 2. If there is no call within this window, we exclude the customer from our analysis because we consider the customer to have churned long before we are able to take pro-active retention measures. We then look 14 days back from the last call to define the end of the predictor window. We used a 30-day predictor window for our analyses here, but it is conceivable to vary this time frame to yield improved results. Note that the exact dates of the predictor window depend on each customer's usage behavior because we want to use the same protocol to prepare new, unlabeled data for the actual prediction.

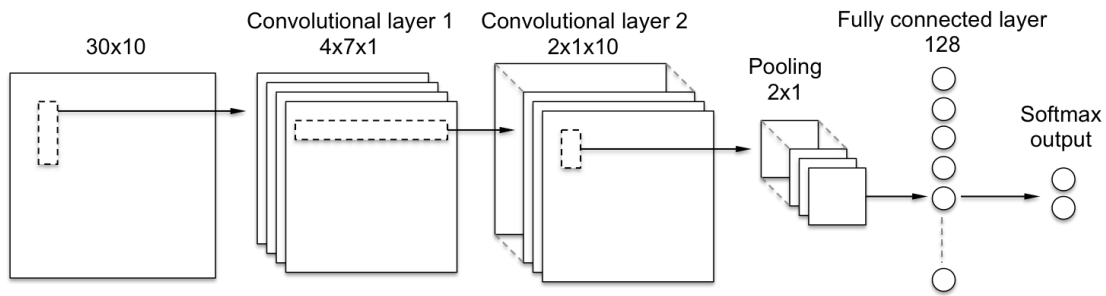

**Figure 3.** One type of architecture used for churn prediction. As shown, a 30 pixel-by-10 pixel input image is used. We refer to this as DL-1. Figures not drawn to scale.

After creating the training and testing images for each customer according to the customer LTL method explained above, we feed them through deep CNNs similar to those used successfully for image classification. One such architecture is shown in Fig. 3, which we call DL-1. This architecture consists of two consecutive convolutional layers, followed by a 2x1 max pooling layer, a fully-connected layer of 128 units, and a softmax output of two units for the binary classification. The first convolutional layer involves four filters of size 7x1, which pans across each usage behavior column over a period of seven days. We chose seven days to analyze the customers' weekly patterns across each usage behavior type at a time. Each filter maintains its shared weights and biases throughout the convolution as commonly employed in image processing. The outputs are then convoluted further in the second convolutional layer, where two filters of size 1x10 pan across all usage behavior features and one row of output from the first convolutional layer. This filter is intended to analyze the customers' usage across all variables at a given time.

After the convolutions, a max pooling layer of size 2x1 is applied that is intended to assist with translational invariance [4]. Next, the fully-connected layer flattens and prepares the data for the softmax output binary classifier. Training and testing this architecture end-to-end yields results superior to that of a CHAID decision tree model when judging by the area-under-the-curve (AUC) benchmark (Table 1). The AUC of a receiver operating curve is a commonly accepted benchmark for comparing models; it accounts for both true and false positives [5,6]. Note that DL-1 was trained for 20 epochs using a binary cross-entropy loss function [7], rectified linear unit activation functions, and stochastic gradient descent by backpropagation [8] in batch sizes of 1000 with adaptive learning rates [9]. Comparing the SPSS CHAID model and the DL-1 model, we see that although both cases exhibit overfitting, the deep learning implementation is superior in both training and testing.

We tested various deep learning hyperparameters and architectures and found the best results in DL-2. DL-2 includes two more features, topup count/amount, and comprises of a 12x7x1 convolutional layer with 0.25 dropout [10], followed by a 2x1 max pooling layer, a 7x1x12 convolutional layer, a 2x1 max pooling layer, a fully-connected layer of 100 units with 0.2 dropout, a fully-connected layer of 40 units with 0.2 dropout, a fully-connected layer of 20 units with 0.2 dropout, and a softmax output of two units for the binary classification. The use of more fully connected layers and dropout in DL-2 appears to reduce overfitting, as evident in the DL-2 AUCs for training and testing datasets in Table 1. While the training AUC is less than that of DL-1, the test AUC is significantly higher. Note that even though 40 epochs were used in DL-2, at 20 epochs it was still superior to DL-1. All other parameters are identical to that of DL-1.

**Table 1.** Training and test AUCs for churn prediction models using SPSS CHAID, DL-1 and DL-2.

| AUC | Training | Test |
|---|---|---|
| **SPSS CHAID** | 0.699 | 0.665 |
| **DL-1** | 0.751 | 0.706 |
| **DL-2** | 0.748 | 0.743 |

So far, we have discussed supervised learning in order to predict churn. To understand customer behavioral patterns and to elucidate the reasons for churning, we can apply unsupervised learning approaches such as autoencoders. Autoencoders are neural networks where the inputs and outputs are identical. They can be used for dimensionality reduction on data and have performed better than principal components analysis [11].

After training an autoencoder with the same dataset used previously, we can produce images that maximally activate the hidden units to obtain the dimensionally-reduced information. If we assume that the input is norm constrained by

$$||x||^2 = \sum_{i=1}^{100} x_i^2 \leq 1,$$  **(Equation 1)**

where $x_i$ is the input to the $i^{th}$ hidden unit, the desired image pixel $x_j$ can be produced from the weights $W_{ij}$ according to [12]

$$x_j = \frac{W_{ij}^{(1)}}{\sqrt{\sum_{j=1}^{100}(W_{ij}^{(1)})^2}}.$$

(Equation 2)

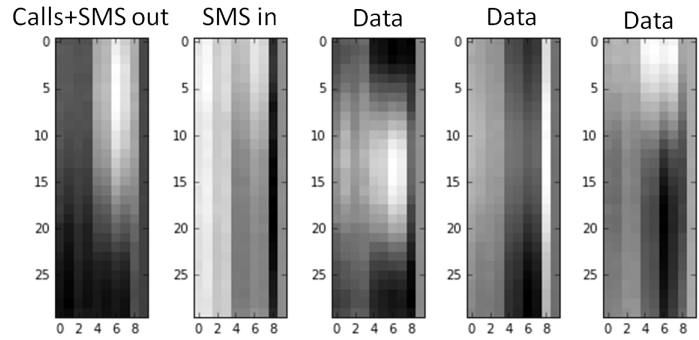

**Figure 4.** Images that maximally activate the hidden units of an autoencoder for the entire customer base. The pixel values increase from white to black.

Such images for the entire customer base are shown in Fig. 4. For this image set, columns 0-3 represent voice calls incoming/outgoing frequency/duration, columns 4-7 represent data download/upload volume/duration, and columns 8-9 represent SMS in/out. One may interpret each real customer's image as being approximately reconstructable from a linear superposition of these base images. It is evident from the second base image that daily incoming marketing SMS messages (solicited and otherwise) are a primary component of all customers. The three different base images regarding data suggest that data usage varies the most among customers and therefore requires the most components to represent.

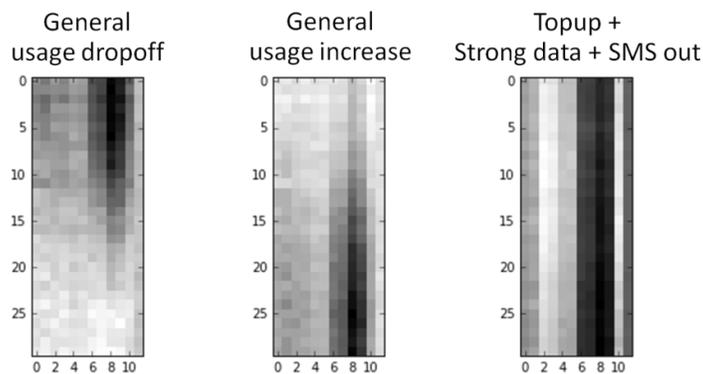

**Figure 5.** Images that maximally activate the hidden units of an autoencoder for customers who churn. The pixel values increase from white to black.

As we are interested in preventing customer churn, we can train an autoencoder on the subset of customers that churn. As shown in Fig. 5, we found three distinguishing base images for customers who churn. In this case, columns 0-1 represent topup frequency/amount, columns 2-5 represent voice calls incoming/outgoing frequency/duration, columns 6-9 represent data download/upload volume/duration, and columns 10-11 represent SMS in/out. The first two images reflect topup, voice and data usage variations throughout the 30-day period.

The third image suggests that many customers who churn have consistent topup, data usage and SMS outgoing throughout the entire 30-day period, but also that they have low voice usage. This leads us to believe that many customers are simply abandoning their SIM because they are not socially tied to the phone number. One possible action that may prevent their churn is to offer voice incentives or promotions. It is also possible that these customers are not receiving adequate SMS marketing messages to maintain their activity in the service.

**Conclusion:** Deep convolutional neural networks and autoencoders prove useful for predicting and understanding churn in the telecommunications industry, outperforming other simpler models such as decision tree modeling. Since no more than 12 temporal features were used for each customer, the input images can be further developed and augmented with more features to improve their efficacy. Another strategy to improve the AUC is to pre-train the weights of the deep convolutional neural network using stacked convolutional autoencoders [13]. We have demonstrated with a more complex type of model (not discussed here) involving of thousands of variables that an AUC of 0.778 is possible. As churn is an important problem to address in many other industries, such as Internet- and subscriptions-based services, we expect that our approach will be widely applicable and adopted in ways beyond what we have covered here.

**Further implementation details:** The deep learning computations were performed on a Dell PowerEdge R630 with Ubuntu 14.04 LTS operating system installed. Docker was used to deploy various systems for development. Computations were performed with open-source libraries, including Theano [14,15], TensorFlow [16] and Keras [17]. The training and testing dataset together consists of over 6 million customers, which are randomly split 80:20, respectively. The churn rate of 3.57% is consistent across all datasets.

**Contributions:** True Corporation provided all the data and hardware. A.W. set up the hardware and software, conceived DL-2 and the autoencoder models, scaled the experiments for 6 million customers, and composed the manuscript. C.B. set up the software and conceived the customer as an image approach, the SPSS CHAID model and DL-1 for initial testing. A.W., C.B., O.L., and R.P. all contributed ideas and reviewed the manuscript.

**Acknowledgements:** A.W. thanks True Corporation for supporting the work as well as Dr. Ian Goodfellow for his endorsement in publishing this article.